%% file: root.tex
\documentclass[letterpaper, 10 pt, conference]{ieeeconf}  % Comment this line out if you need a4paper

\IEEEoverridecommandlockouts                              % This command is only needed if 
\input{preamble-commands}

\title{\LARGE \bf
Real-Time Ellipse Detection for Robotics Applications}

\author{Azarakhsh Keipour$^{1}$, 
Guilherme A. S. Pereira$^{2}$ 
and Sebastian Scherer$^{3}$% <-this % stops a space
\thanks{$^{1,3}$ Robotics Institute, Carnegie Mellon University, Pittsburgh, PA
        {\tt\small [keipour, basti]@cmu.edu}}%
\thanks{$^{2}$ Department of Mechanical and Aerospace Engineering, West Virginia University, WV 
        {\tt\small guilherme.pereira@mail.wvu.edu}}%
}

\begin{document}

\maketitle
\thispagestyle{empty}
\pagestyle{empty}

%%%%%%%%%%%%%%%%%%%%%%%%%%%%%%%%%%%%%%%%%%%%%%%%%%%%%%%%%%%%%%%%%%%%%%%%%%%%%%%%
\begin{abstract}

\input{abstract}

\end{abstract}

%%%%%%%%%%%%%%%%%%%%%%%%%%%%%%%%%%%%%%%%%%%%%%%%%%%%%%%%%%%%%%%%%%%%%%%%%%%%%%%%

\input{1.Intro}
\input{2.Approach}
\input{3.Experiments}
\input{4.Conclusion}

\addtolength{\textheight}{0.0cm}   % This command serves to balance the column lengths

%%%%%%%%%%%%%%%%%%%%%%%%%%%%%%%%%%%%%%%%%%%%%%%%%%%%%%%%%%%%%%%%%%%%%%%%%%%%%%%%

%%%%%%%%%%%%%%%%%%%%%%%%%%%%%%%%%%%%%%%%%%%%%%%%%%%%%%%%%%%%%%%%%%%%%%%%%%%%%%%%

%%%%%%%%%%%%%%%%%%%%%%%%%%%%%%%%%%%%%%%%%%%%%%%%%%%%%%%%%%%%%%%%%%%%%%%%%%%%%%%%
\section*{ACKNOWLEDGMENT}

Authors want to thank Near Earth Autonomy for their support in this project. The project was sponsored by Carnegie Mellon University Robotics Institute and Mohamed Bin Zayed International Robotics Challenge. During the realization of this work, G. Pereira was supported by UFMG and CNPq/Brazil.

%%%%%%%%%%%%%%%%%%%%%%%%%%%%%%%%%%%%%%%%%%%%%%%%%%%%%%%%%%%%%%%%%%%%%%%%%%%%%%%%

\bibliographystyle{IEEEtran}
%\addbibresource{paper-citations.bib}
\bibliography{paper-citations.bib}

\end{document}

%% file: preamble-commands.tex
\usepackage{amssymb,amsmath,amsfonts}
\usepackage{graphicx}
\usepackage[table,xcdraw,dvipsnames]{xcolor}
\usepackage{tikz}
\usepackage{url}

\usepackage{subcaption}
\usepackage{colortbl}
\usepackage{booktabs}
\usepackage{array}
\usepackage{tabularx}

\usepackage[flushleft]{threeparttable}
\usepackage{multirow}

\let\labelindent\relax
\usepackage{enumitem}

\newcolumntype{L}[1]{>{\raggedright\let\newline\\\arraybackslash\hspace{0pt}}m{#1}}

\makeatletter
\DeclareRobustCommand\onedot{\futurelet\@let@token\@onedot}
\def\@onedot{\ifx\@let@token.\else.\null\fi\xspace}

\usepackage[T1]{fontenc}  % For correct {}s in \texttt
\usepackage[b]{esvect}    % For \vv

\makeatletter
\newlength\xvec@height%
\newlength\xvec@depth%
\newlength\xvec@width%
\newcommand{\xvec}[2][]{%
  \ifmmode%
    \settoheight{\xvec@height}{$#2$}%
    \settodepth{\xvec@depth}{$#2$}%
    \settowidth{\xvec@width}{$#2$}%
  \else%
    \settoheight{\xvec@height}{#2}%
    \settodepth{\xvec@depth}{#2}%
    \settowidth{\xvec@width}{#2}%
  \fi%
  \def\xvec@arg{#1}%
  \def\xvec@dd{:}%
  \def\xvec@d{.}%
  \raisebox{.2ex}{\raisebox{\xvec@height}{\rlap{%
    \kern.05em%  (Because left edge of drawing is at .05em)
    \begin{tikzpicture}[scale=1]
    \pgfsetroundcap
    \draw (.05em,0)--(\xvec@width-.05em,0);
    \draw (\xvec@width-.05em,0)--(\xvec@width-.15em, .075em);
    \draw (\xvec@width-.05em,0)--(\xvec@width-.15em,-.075em);
    \ifx\xvec@arg\xvec@d%
      \fill(\xvec@width*.45,.5ex) circle (.5pt);%
    \else\ifx\xvec@arg\xvec@dd%
      \fill(\xvec@width*.30,.5ex) circle (.5pt);%
      \fill(\xvec@width*.65,.5ex) circle (.5pt);%
    \fi\fi%
    \end{tikzpicture}%
  }}}%
  #2%
}
\makeatother

\makeatletter
\renewcommand*\env@matrix[1][\arraystretch]{%
  \edef\arraystretch{#1}%
  \hskip -\arraycolsep
  \let\@ifnextchar\new@ifnextchar
  \array{*\c@MaxMatrixCols c}}
\makeatother

\definecolor{commentcolor}{gray}{0.5}
\usepackage{algorithm}
\usepackage{algpseudocode}
\algrenewcommand\algorithmicindent{1.0em}%

\algnewcommand{\LineComment}[1]{\State \textcolor{commentcolor}{\(\triangleright\) #1}}
\algnewcommand{\To}{\textbf{to}}
\algnewcommand{\Break}{\textbf{break}}
\algnewcommand{\Continue}{\textbf{continue}}
\algnewcommand{\IIf}[1]{\State\algorithmicif\ #1\ \algorithmicthen}
\algnewcommand{\EndIIf}{\unskip}
\algnewcommand{\var}[1]{\textit{#1}}
\algnewcommand{\func}[1]{\textsc{#1}}

%% file: abstract.tex
We propose a new algorithm for real-time detection and tracking of elliptic patterns suitable for real-world robotics applications. The method fits ellipses to each contour in the image frame and rejects ellipses that do not yield a good fit. The resulting detection and tracking method is lightweight enough to be used on robots' resource-limited onboard computers, can deal with lighting variations and detect the pattern even when the view is partial. The method is tested on an example application of an autonomous UAV landing on a fast-moving vehicle to show its performance indoors, outdoors, and in simulation on a real-world robotics task. The comparison with other well-known ellipse detection methods shows that our proposed algorithm outperforms other methods with the F1 score of 0.981 on a dataset with over 1500 frames. The videos of experiments, the source codes, and the collected dataset are provided with the paper.

%% file: 1.Intro.tex
\section{INTRODUCTION} \label{sec:intro}

Real-time detection and tracking of circular and elliptic shapes using an onboard vision system are essential for many real-world (mainly robotics) applications. For example, many aerial robot' landing zones consist of an elliptical shape~\cite{cargozone}, and autonomous cars need to detect the circular road signs~\cite{roadsign}. 

Detecting ellipses in images has been a topic of interest for researchers for a long time \cite{tsuji1977detection, tsuji1978detection}. In general, it is possible to classify the available ellipse detection approaches into four classes. 

The first class contains voting-based algorithms, including Hough Transform (HT) \cite{cite:hough} and the methods based on it. The HT algorithm uses a 5-dimensional parametric space for the ellipse detection task and is too slow for real-time applications. Other methods try to enhance HT by reducing the dimensionality in parametric space \cite{cite:houghdim1, cite:houghdim2, cite:zhang}, performing piecewise-linear approximation for the curved segments \cite{cite:houghapprox1} or randomizing the method \cite{cite:houghrand1, cite:houghrand2, cite:houghrand3}. Some of these modified HT-based methods are fast but generally less accurate and unsuitable for many robotics applications.

The second class contains optimization-based methods. Most popular methods convert the ellipse fitting problem into a least-squares optimization problem and solve the problem to find the best fit \cite{cite:fitls, fitzgibbon1995buyer}. These methods are generally not robust and tend to produce many false positives. However, their high processing speed is useful for quickly estimating the ellipse as the first step in other methods. Another group of optimization-based methods try to solve the non-linear optimization problem of fitting an ellipse using a genetic algorithm \cite{cite:genetic1, cite:genetic2}. These algorithms perform well in noisy images with multiple ellipses, but their processing time makes them impractical for real-time applications.

The third class consists of methods based on edge linking and curve grouping \cite{cite:mai}. These methods can detect ellipses from complex and noisy images but are generally computationally expensive and cannot be used in real-time applications.

The last class contains the methods that use an ad-hoc approach or combine the methods from the first three classes to achieve a faster and more accurate ellipse detection. A real-time method proposed by Nguyen et al. \cite{cite:nguyen} detects arc-segments from the image and groups them into elliptical arcs to estimate the ellipse parameters using a least-square optimization. A method proposed by Fornaciari et al.~\cite{fornaciari2014ellipsedetect} combines arc grouping with Hough Transform in a decomposed parameter space to estimate the ellipse parameters. In this way, it achieves a real-time performance comparable to slower, more robust algorithms.

Currently, in many robotics applications, the detection of the previously-known elliptic objects is done by general learning-based object detection methods~\cite{bhattacharya2021}. However, this approach requires collection and enhancement of large amount of data for each individual object and access to powerful computing systems for training. Recently, there has been an effort to implement a learning-based method specifically for ellipse detection~\cite{cnnellipse}. The method has been shown to work well for occlusions and partial views. However, it requires a powerful processor and is not yet suitable for real-time execution on onboard computers.

While many ellipse detection algorithms are proposed for computer vision tasks (\cite{WANG2021107741, Liu2020, 8972900}), the performance of these methods drops when used in real-world robotics tasks. Some of the challenges in these applications include:

\begin{itemize}[leftmargin=*]
\item The algorithm should work online (with a frequency greater than 10 Hz), generally on the robot's resource-limited onboard computer.

\item The elliptical object should be detected when it is far from or close to the robot.

\item The shape of the ellipse is transformed by a projective distortion, which occurs when the shape is seen from different views.

\item There is a wide range of possible illumination conditions (e.g., cloudy, sunny, morning, evening, indoor lighting). 

\item Due to reflections (e.g., from sources like sun, bulbs), the pattern may not always be seen in all the frames, even when the camera is close.

\item In some frames, there may be shadows on the pattern (e.g., the shadow of the robot or trees around).

\item In some frames, the pattern may be seen only partially (due to occlusion or being outside of the camera view).

\item In some frames, the pattern may not be present.

\end{itemize}

Different approaches have been devised in the literature to detect and track an elliptic pattern in real applications to address the challenges above. For example, \cite{beul2018} uses the circular Hough-transform algorithm for the initial detection, which is a slow algorithm, and then uses other features in their pattern for the tracking. The authors of~\cite{jin2018} have developed a Convolutional Neural Network trained with over 5,000 images collected from the pattern moving at a maximum speed of $15$\,km/h at various heights to select (validate) the desired ellipse from all the ellipses detected using the  method in~\cite{jia2017}. The authors of~\cite{li2018} use the  method proposed in~\cite{fornaciari2014ellipsedetect} for ellipse detection only when their robot is far from the pattern and switch to other features in their pattern for the closer frames. 

It is possible to use a predictor filter along with the detector to narrow down the region of interest (ROI) and improve the detection speed of ellipses across the frames. The authors of~\cite{ellipsekalman1} and \cite{ellipsekalman2} use Kalman filter to predict the direction of the ellipse movement in the frames, increase the detection speed and reduce the false positives. In many robotics applications, the combined simultaneous motions of the elliptic pattern, the robot, and the camera complicate the prediction provided by such predictors, resulting in the loss of tracking on many frames, triggering a detection on the entire frame and slowing the tracking speed.

This paper proposes a novel ellipse detection method that performs on resource-limited onboard computers in real-time. Our proposed ad-hoc method first extracts all contours from the input image and then uses a least-square method to fit an ellipse to each contour. It tests how well the estimated ellipse fits the contour and starts rejecting the contours using several criteria. The remaining contours are accepted as resulted ellipses, and if there are no contours left (e.g., due to the absence of ellipses in the frame), the algorithm returns an empty set of ellipses. The detection method is combined with a simple tracking algorithm that changes detection parameters as necessary and significantly increases the elliptical pattern detection's precision and performance. This simple tracker has been shown to be faster than using predictor filters due to the reduced tracking losses. The resulting detector and tracking method can deal with lighting variations and detect the pattern even when the view is partial. By comparing the results of our experiments on a collected dataset to the other methods tested on similar datasets we show that our method outperforms all the other real-time ellipse detection methods proposed so far.

Section~\ref{sec:approach} explains the developed method for detection and tracking of the elliptical targets; Section~\ref{sec:tests} shows an example real-time application that uses the proposed algorithm and compares the performance with some other available well-known methods. Finally, Section~\ref{sec:conclusion} discusses how the proposed method can be further improved in the future.

%% file: 2.Approach.tex
\section{ELLIPSE DETECTION AND TRACKING} \label{sec:approach}

The idea of the proposed ellipse detection algorithm is to fit ellipses to all the contours in a frame or the region of interest and then decide if the ellipse is a good fit. With real-time ellipse fitting algorithms and fast criteria for checking the fit, the result is a real-time detection algorithm that can detect ellipses as long as the elliptic pattern's contours are (at least partially) extracted during the contour extraction. With a suitable contour extraction method, the whole detection becomes robust to the lighting and environmental changes. 

The pseudocode for the proposed ellipse detector is shown in Algorithm~\ref{alg:detection}.
\begin{algorithm}[!t]
\caption{Proposed approach for ellipse detection}
\label{alg:detection}
\begin{algorithmic}[1]
\LineComment {This function detects all the ellipses found using the input criteria (thresholds)}
\Function{DetectEllipses}{\textit{frame}, \textit{thresholds}}
    \LineComment {Initialize an empty set for the detected ellipses}
    \State $\textit{Detections} \gets \text{$\emptyset$}$
    \State     
    \LineComment {Detect all edges in the frame}
    \State $\textit{edges} \gets \text{\textsc{DetectEdges}(\textit{frame})}$
    \State     
    \LineComment {Extract contours from the detected edges}
    \State $\textit{contours} \gets \text{\textsc{ExtractContours}(\textit{edges})}$
    \State     
    \ForAll {$\textit{c} \in \textit{contours}$}
        \LineComment {Reject if the contour is too small}
        \IIf {$|c.pixels| < minContourSize$} \Continue \EndIIf
        \State     
        \LineComment {Fit an ellipse to each contour}
        \State $\textit{ellipse} \gets \text{\textsc{FitEllipse}(\textit{c})}$
        \State     
        \LineComment {Reject ellipses with unreasonable dimensions}
        \IIf {$ellipse.largeAxis > mxAxSize$} \Continue \EndIIf
        \IIf {$ellipse.smallAxis < mnAxSize$} \Continue \EndIIf
        \State $\textit{axisRatio} \gets \textit{ellipse.largeAxis / ellipse.smallAxis}$
        \IIf {$axisRatio > maxAxisRatio$} \Continue \EndIIf
        \State     
        \LineComment {Reject contours with small overlap with ellipse}
        \State $\textit{overlap} \gets \textit{$c.pixels  \  \cap \  ellipse.pixels$}$
        \IIf {$|\textit{overlap}| \textit{ / } |\textit{c.pixels}| \text{ is small}$} \Continue \EndIIf
        \State     
        \LineComment {Reject ellipses with small overlap with the edges}
        \State $\textit{overlap} \gets \textit{$ellipse.pixels \  \cap \  edges.pixels$}$
        \IIf {$|\textit{overlap}| \textit{ / } |\textit{ellipse.pixels}| \text{ is small}$} \Continue \EndIIf
        \State     
        \LineComment {Add the detected ellipse to the set of detections}
        \State $\textit{Detections} \text{.Insert(\textit{ellipse})}$
    \EndFor
    \State     
    \LineComment {Return the detections after all contours are processed}
    \State \Return $\textit{Detections}$
\EndFunction
\end{algorithmic}
\end{algorithm}
The detection function receives a frame and a set of threshold values used in the function and returns a set of detected ellipses. A step-by-step explanation of the algorithm is as follows:

\begin{enumerate}[leftmargin=*]
\item The first step of the algorithm is to extract edges from the input frame and create the edge image. We used the Canny edge detector \cite{canny1986computational}, considering that the thresholds should be selected carefully to extract suitable edges in a large variety of conditions (e.g., illumination) while preventing the generation of too many edges. Usually, it is beneficial for smaller targets to produce more edges; this action will increase the processing time but reduces the probability of not having an edge for the elliptic target.

\item The resulted edges are utilized to extract contours using the algorithm proposed by Suzuki and Abe~\cite{suzuki1985topological}. This step helps make connections between relevant edges and enables the extraction of shapes in a frame. Ideally, each contour is a collection of connected points constructing a shape's border in the edge image.

\item Each contour is processed individually to determine if it is a part of an ellipse or not. For robotics tasks, an ellipse can have one of the following contour types:

\begin{figure}[t]
\centering
\includegraphics[width=\linewidth]{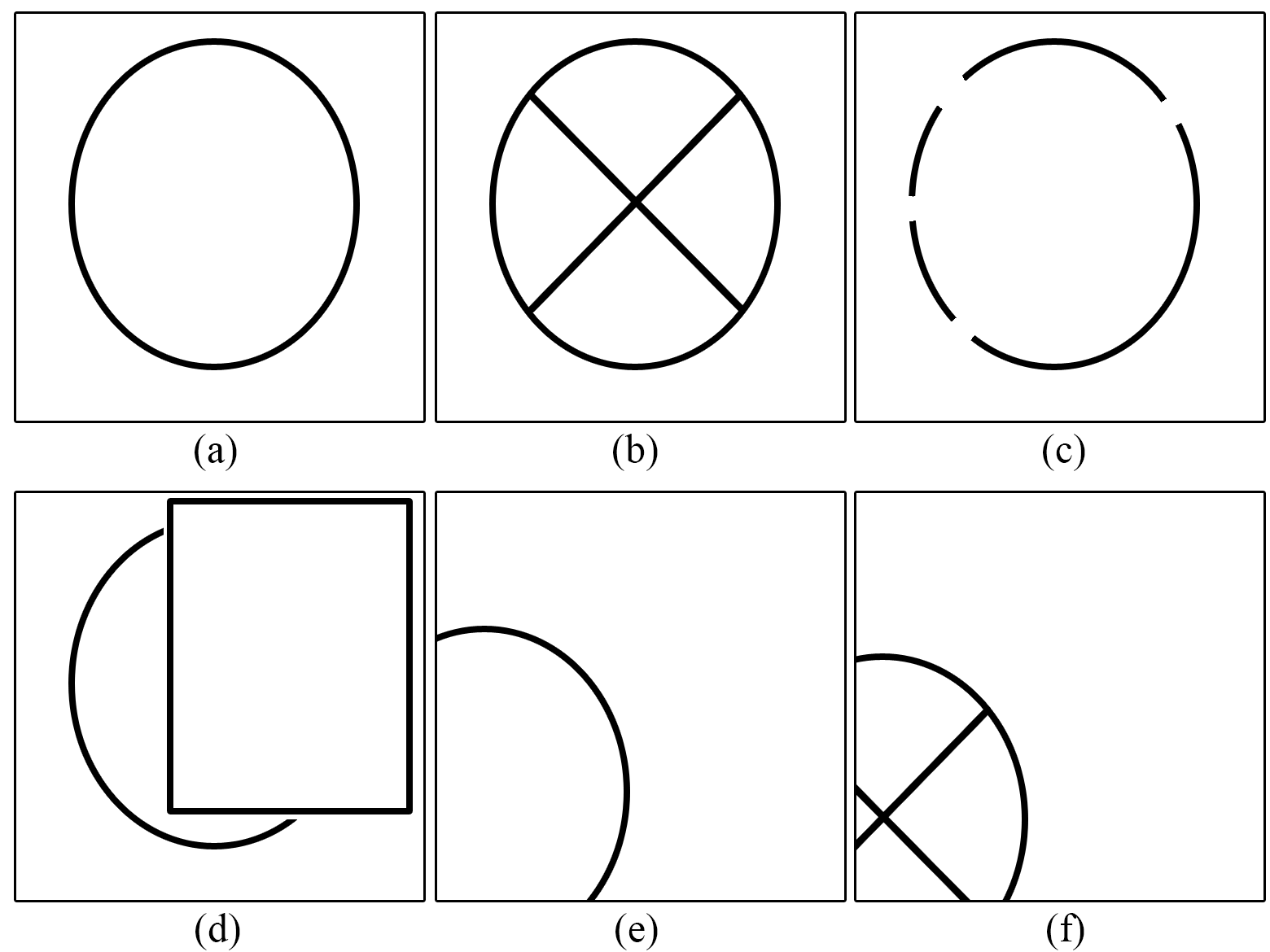}
\caption{Contour types of ellipses in different conditions: (a) A single contour containing a full ellipse. (b) A single contour containing a full ellipse with additional connected contours from the rest of the pattern. (c) Multiple contours, together constructing a full ellipse. (d) A single contour containing an ellipse that is partially occluded by another object. (e) A single contour containing a partially-seen ellipse. (f) A single contour containing a full ellipse with additional connected contour branches that is partially seen in the frame.}
\label{fig:circletypes}
\end{figure}

    \begin{itemize}[leftmargin=*]
    \item A single contour containing a full ellipse without any occlusions, broken segments or additional contours (Fig.~\ref{fig:circletypes}(a)).
    \item A single contour containing an ellipse with additional connected contours from the rest of the pattern (Fig.~\ref{fig:circletypes}(b)).
    \item Multiple contours, together constructing a full ellipse (Fig.~\ref{fig:circletypes}(c)).
    \item A single contour containing an ellipse that is partially occluded by other objects (Fig.~\ref{fig:circletypes}(d)).
    \item A single contour containing an ellipse that is partially seen in the frame (Fig.~\ref{fig:circletypes}(e)).
    \item A combination of the above contour types (Fig.~\ref{fig:circletypes}(f)).
    \end{itemize}

In order to correctly detect the above contour types, the following process is performed on each contour:

\begin{enumerate}[leftmargin=*]
  \item If a contour has a very small number of pixels, it is ignored since it is most probably just noise. 
  
  \item An ellipse is fit to the contour using the least-square approximation method described by Fitzgibbon and Fisher \cite{fitzgibbon1995buyer}. The method fits an ellipse to any input contour; therefore, the contour should be further processed to determine the actual ellipses. 
  
  \item The resulting ellipse will be ignored if any of the axes are too small or if the ellipse's eccentricity is high (close to 1). In high eccentricity, the resulting ellipse is similar to a line and is not really an ellipse.

  \item The current contour is intersected with the perimeter of the resulted ellipse, described by its center point, minor and major axes, and rotation angle. Then, the ratio of the number of pixels in the intersection to the number of all pixels in the contour is calculated:
  \begin{equation}
    ContourOverlap = \frac{|Contour \  \cap \  Ellipse|}{|Contour|},
    \label{eq:01}
  \end{equation}
  
 where $| \cdot |$ is used for the number of pixels. A low result means that the contour and the resulted ellipse do not fit well, and a significant portion of the contour is not lying on the fitted ellipse. In this case, the contour is ignored and not further processed.
  
  \item Finally, the ellipse is intersected with the edges, and the ratio of the number of pixels in the intersection to the number of all the pixels in the ellipse is calculated:
  \begin{equation}
    EllipseOverlap = \frac{|Ellipse \  \cap \  Edges|}{|Ellipse|}, 
    \label{eq:02}
  \end{equation}
  
where $| \cdot |$ is used for the number of pixels. A low result means that a significant portion of the ellipse does not correspond to any contours in the image. The reason that the ellipse is intersected with the edge image instead of only its constructing contour is that due to noise or imperfect contour detection step, sometimes an ellipse is broken into two or more contours (e.g., the cases like Fig. \ref{fig:circletypes}(c)). In these cases, checking the ellipse against a single contour will give a low ratio and results in false negatives. To take care of the cases similar to Figure~\ref{fig:circletypes}(e), it is essential to count only the ellipse pixels that are actually lying in the image; otherwise, the result will be too low, and a partially viewed ellipse may get rejected. Additionally, to make the algorithm more robust to slightly imperfect ellipses, increasing the edges' thickness before intersecting them with the ellipse is beneficial. 
  
  \item If an ellipse is not rejected in the previous steps, it represents a real ellipse in the image and is added to the set of detected ellipses.
\end{enumerate}

\item Due to the target ellipse's thickness, there is a chance of detecting two or more concentric ellipses. Therefore, the returned set of detected ellipses in the frame is further processed to find the concentric ellipses. All the non-concentric sets of ellipses are ignored when this happens.

\end{enumerate}

The proposed algorithm can also detect the ellipses with partial occlusion or the ellipses exceeding the image boundaries. The rejection criteria can be chosen in a way to accept ellipses in such cases. 

Choosing higher rejection criteria for ellipse detection generally helps to eliminate potential false positives. Therefore, it is beneficial to have higher rejection thresholds when there is no prior information about the pattern location and size in the image. However, after the first detection of the elliptical pattern, it is possible to change the initial parameters and conditions for the detection to enhance the performance and increase the detection rate. 
Decreasing the detection threshold values reduces the probability of losing the target ellipse in the following frames due to illumination variation, noise, occlusion, or other changes in the conditions. Additionally, if the approximate movement of the target is known, setting the region of interest (ROI) to the area where the pattern is expected to be seen will decrease both the processing time and the probability of falsely detecting other similar shapes that were rejected in the initial frame due to the higher thresholds.
Furthermore, whenever the detected target is large enough to be detected in the input frame with a smaller scale, the frame can be scaled down to increase the processing speed. Performing ellipse detection on the smaller frame takes less CPU time and is much faster. 

We propose these steps that can be combined with the ellipse detection algorithm to track the detected elliptic pattern in the next frames more efficiently:

\begin{enumerate}[leftmargin=*]
\item Significantly decrease the $ContourOverlap$ and $EllipseOverlap$ threshold values. This threshold change increases the detection rate and helps the algorithm to keep tracking the target.
\item Determine the region of interest, which includes the current detected target and expands in all directions based on the distance from the target, the robot's relative speed and the pattern, and other available information. For example, if the distance is far and the relative linear and angular speeds are low, the target is expected to be found close to the current detection coordinates in the next frame. 
\item Decreasing the scale of the frame by order of two (up until a set threshold) every time that the detected target is larger than a specified size and scaling the frame up again by order of two (up to the actual frame scale) every time the detected target is smaller than a chosen threshold. To make the approach more robust, ellipse detection with initial higher parameters is performed once again on a higher scale if no candidate targets are found on a lower scale. The scale change is performed to reduce the execution time, as the algorithm needs to process only a quarter number of pixels every time it scales the frame down.
\end{enumerate}

\begin{algorithm}
\caption{Proposed approach for elliptic target tracking}
\label{alg:tracking}
\begin{algorithmic}[1]
\LineComment {This function reads the frames from a video stream and performs tracking of a target}
\Function{TrackTarget}{\textit{videoStream}}

    \LineComment {Set frame scale, ROI and tracking status}
    \State \textit{isTracking} $\gets$ \textbf{false}
    \State \textit{scale} $\gets$ 1
    \State \textit{roi} $\gets$ $\emptyset$
	\State
	\While {videoStream $\neq$ $\emptyset$}
%        \LineComment {Read a video frame from video stream}
        \State \textit{frame} $\gets$ \textsc{ReadFrame}(\textit{videoStream})

        \LineComment {Detect the ellipse in the frame or ROI}
        \If {\textit{isTracking} $=$ \textbf{false}}
        	\State \textit{offsetTarget} $\gets$ \textsc{DetectTarget}(\textit{frame}, \textit{scale})
        \Else
            \State \textit{offsetTarget} $\gets$ \textsc{DetectTarget}(\textit{roi}, \textit{scale})
        \EndIf

    	\State
        \If {\textit{offsetTarget} $=$ $\emptyset$ \textbf{and} \textit{scale} $>$ 1}
        	\LineComment {If target not found, try on higher frame scale}
        	\State \textit{scale} $\gets$ \textit{scale} $\div$ 2
            \If {\textit{isTracking} $=$ \textbf{false}}
            	\State \textit{offsetTarget} $\gets$ \textsc{DetectTarget}(\textit{frame}, \textit{scale})
            \Else
            	\State \textit{offsetTarget} $\gets$ \textsc{DetectTarget}(\textit{roi}, \textit{scale})
            \EndIf
        \Else
        	\LineComment {If found, set the proper scale for next frame}
            \State \textit{isTracking} $\gets$ \textbf{true}
            \If {\textit{offsetTarget.size} $>$ \textit{maxTargetSize}}
                \If {\textit{scale} $<$ \textit{maxScale}}
            	    \State \textit{scale} $\gets$ \textit{scale} $\times$ 2
            	\EndIf
            \ElsIf {\textit{offsetTarget.size} $<$ \textit{minTargetSize}}
                \If {\textit{scale} $>$ 1} 
            	    \State \textit{scale} $\gets$ \textit{scale} $\div$ 2
            	\EndIf
            \EndIf
        \EndIf
        \State
    
    	\LineComment {Compensate the target offset resulted from detection in ROI. Do nothing if ROI is null}
        \State \textit{target} $\gets$ \textsc{CompensateOffset}(\textit{offsetTarget}, \textit{roi})
        \State 
    
    	\LineComment {Set the tracking status and ROI for the next frame}
        \If {\textit{target} $=$ $\emptyset$}
        	\State \textit{isTracking} $\gets$ \textbf{false}
        	\State \textit{roi} $\gets$ $\emptyset$
        \Else
        	\State \textit{isTracking} $\gets$ \textbf{true}
            \LineComment {Expand the area around the detected target}
            \State \textit{roi} $\gets$ \textsc{CalculateROI}(\textit{target})
        \EndIf
        \State
        
    	\State \Return $\textit{target}$
	\EndWhile
\EndFunction
\end{algorithmic}
\end{algorithm}

\begin{algorithm}
\caption{Proposed approach for elliptic target detection in a given frame scale}
\label{alg:targetdetection}
\begin{algorithmic}[1]
\LineComment {This function detects the target in the input frame at the specified frame scale}
\Function{DetectTarget}{\textit{frame}, \textit{scale}, \textit{isTracking}}
	\State
    \LineComment {Resize the frame to make the processing faster}
    \State \textit{scaledFrame} $\gets$ \textsc{ScaleImage}(\textit{frame}, 1 / \textit{scale})
    \State
        
    \LineComment {Set detection parameters based on inputs}
    \State \textit{thresholds} $\gets$ \textsc{DetermineParams}(\textit{isTracking}, \textit{scale})
    \State

    \LineComment {Detect ellipses using the specified thresholds}
    \State \textit{detections} $\gets$ \textsc{DetectEllipses}(\textit{scaledFrame}, \textit{thresholds})
    \State

    \LineComment {Select the actual target from the detected ellipses}
    \State \textit{scaledTarget} $\gets$ \textsc{SelectTarget}(\textit{detections})
    \State
	
    \LineComment {Rescale the detected target to original scale}
    \State \textit{target} $\gets$ \textsc{ScaleSize}(\textit{scaledTarget})
    \State

    \LineComment {Return the detected target (or $\emptyset$ if not detected)}
    \State \Return $\textit{target}$
\EndFunction
\end{algorithmic}
\end{algorithm}

Algorithms~\ref{alg:tracking} and~\ref{alg:targetdetection} show the described method in more detail.

%% file: 3.Experiments.tex
\section{EXPERIMENTS AND RESULTS} \label{sec:tests}

\subsection{Elliptic Target Dataset and Parameter Selection} \label{sec:dataset}

We created a dataset with sequences recorded using a UAV from a stationary and moving vehicle carrying an elliptical platform at various distances, angles, and illumination conditions. The dataset contains 1,511 frames (1,378 positive and 133 negative frames) and 456 frames of a thinner version of the same pattern. The size of the frames is $640 \times 360$, and the ground truth for the detections is provided. The dataset can be accessed from \url{http://theairlab.org/landing-on-vehicle}.

The thresholds selection of the proposed ellipse detection algorithm depends on the tolerance for false positives vs. false negatives in the application. However, in practice, the detection is not too sensitive to the parameters in most cases, and they can be selected from a broad range. For our tests on the AirLab Elliptic Target Detection Dataset, we empirically chose the values shown in Table~\ref{tbl:ellipse-detection-parameters}. The GUI tool provided with the code helps with the calibration process letting the user see the parameters' effects in real-time. The ellipse detection parameters are independent of the lighting conditions. Therefore after a one-time calibration, the algorithm should detect the pattern in a wide range of weather conditions (e.g., sunny, cloudy, snowy) as long as the light is enough for the camera to capture the pattern.

\begin{table}[!t]
\centering
\begin{threeparttable}
\caption{Ellipse Detection parameters chosen for the tests on the AirLab Elliptic Target Detection Dataset.}
\label{tbl:ellipse-detection-parameters}
\begin{tabular}{|l|c|l|}
\hline
\rowcolor[HTML]{EFEFEF} 
Parameter   & Value & Justification \\ \hline
\textit{ContourOverlap} for detection & 0.95 & To prevent False Positives. \\ \hline
\textit{ContourOverlap} for tracking  & 0.7 & To enhance target tracking.  \\ \hline
\textit{EllipseOverlap} for detection & 0.95 & To prevent False Positives.  \\ \hline
\textit{EllipseOverlap} for tracking  & 0.3 & To enhance target tracking.  \\ \hline
\end{tabular}
\end{threeparttable}
\end{table}

\begin{figure}[!t]
\centering
    \begin{subfigure}[b]{0.238\textwidth}
        \includegraphics[width=\textwidth]{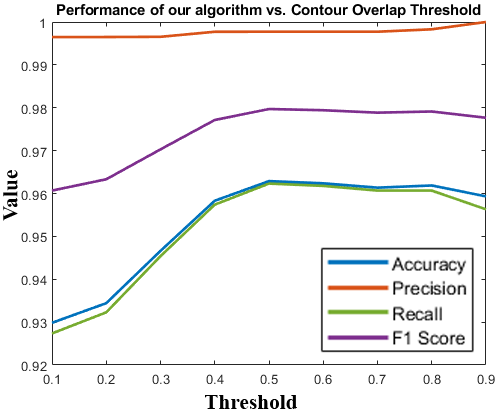}
        \caption{~}
    \end{subfigure}
    \hfill
    \begin{subfigure}[b]{0.238\textwidth}
        \includegraphics[width=\textwidth]{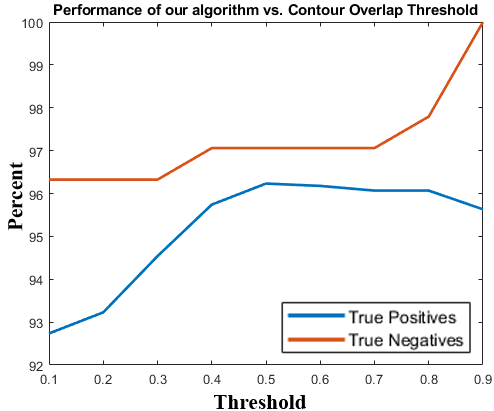}
        \caption{~}
    \end{subfigure}
    
    \medskip
    
    \begin{subfigure}[b]{0.238\textwidth}
        \includegraphics[width=\textwidth]{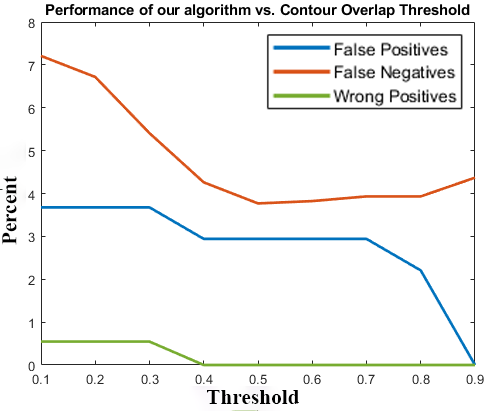}
        \caption{~}
    \end{subfigure}
    \hfill
    \begin{subfigure}[b]{0.238\textwidth}
        \includegraphics[width=\textwidth]{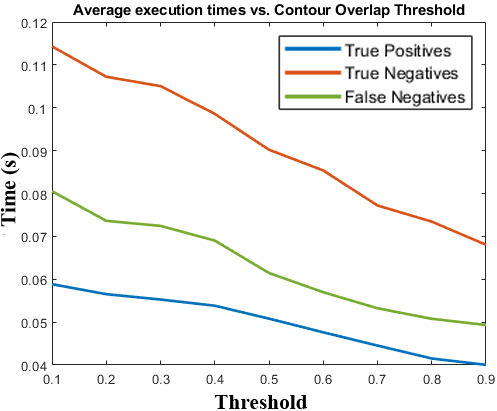}
        \caption{~}
    \end{subfigure}
\caption{Performance and execution times of our algorithm vs. contour overlap threshold (\textit{ContourOverlap} parameter). The value of \textit{EllipseOverlap} is fixed to $0.7$ for all the experiments. Wrong Positives are the wrong detections when the target was present in the frame.
(a) Accuracy, Recall, and F1 Score of the algorithm increase with the increase of \textit{ContourOverlap} up to a point. (b) Increasing \textit{ContourOverlap} increases the number of True Negatives while may result in fewer True Positives after some point. (c) Increasing \textit{ContourOverlap} results in a lower number of False Positives and Wrong Detections, while it may result in an increase of the number of False Negatives after some point. (d) The execution time decreases by increasing the \textit{ContourOverlap} parameter.}
\label{fig:contour-parameter-selection}
\end{figure}

\begin{figure}[!t]
\centering
    \begin{subfigure}[b]{0.238\textwidth}
        \includegraphics[width=\textwidth]{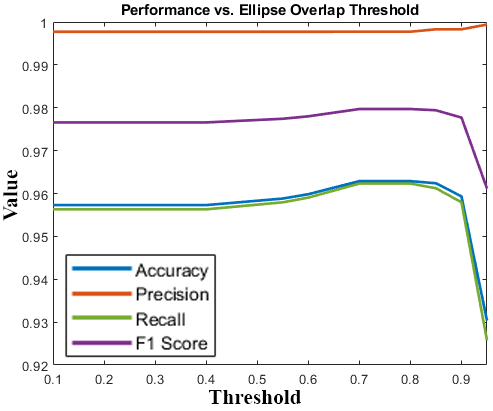}
        \caption{~}
    \end{subfigure}
    \hfill
    \begin{subfigure}[b]{0.238\textwidth}
        \includegraphics[width=\textwidth]{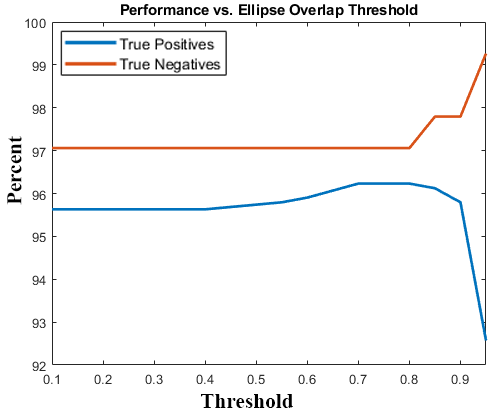}
        \caption{~}
    \end{subfigure}
    
    \medskip
    
    \begin{subfigure}[b]{0.238\textwidth}
        \includegraphics[width=\textwidth]{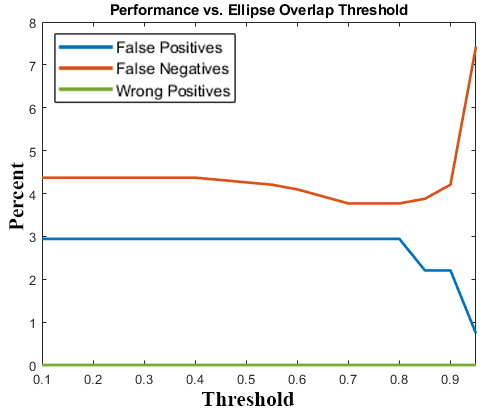}
        \caption{~}
    \end{subfigure}
    \hfill
    \begin{subfigure}[b]{0.238\textwidth}
        \includegraphics[width=\textwidth]{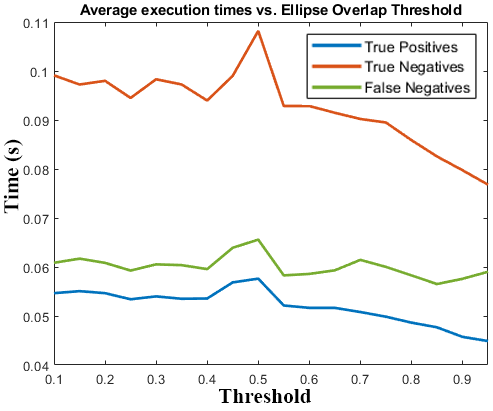}
        \caption{~}
    \end{subfigure}
\caption{Performance of our algorithm vs. ellipse overlap threshold (\textit{EllipseOverlap} parameter). The value of \textit{ContourOverlap} is fixed to $0.5$ for all the experiments.
(a) Accuracy, Recall, and F1 Score of the algorithm very slowly increase with the increase of \textit{EllipseOverlap} up to a breaking point, where they suddenly drop. (b) Increasing \textit{EllipseOverlap} increases the number of True Negatives while reducing the number of True Positives after a certain point. (c) Increasing \textit{EllipseOverlap} results in a lower number of False Positives, while it may result in an increase of the number of False Negatives after some point. (d) The execution time generally decreases by increasing the \textit{EllipseOverlap} value.}
\label{fig:ellipse-parameter-selection}
\end{figure}

In order to assess the sensitivity of the algorithm against the thresholds, Figure~\ref{fig:contour-parameter-selection} shows the performance of the algorithm for different values of \textit{ContourOverlap} with the value of \textit{EllipseOverlap} fixed to $0.7$. Additionally, Figure~\ref{fig:ellipse-parameter-selection} shows the performance of the algorithm for different values of \textit{EllipseOverlap} with the value of \textit{ContourOverlap} fixed to $0.5$. 

Let us define $N_{TP}$, $N_{FP}$, $N_{TN}$, $N_{FN}$, $N_{All}$, and $N_{WD}$ as the number of True Positive detections, False Positives, True Negatives, False Negatives, the total number of frames and the number of frames with visible targets where the detection was wrong. The measures in the plots are defined as follows using the five outcome types above.

\begin{itemize}[leftmargin=*]
\item Accuracy measures the ratio of all the frames that the algorithm gives the correct result (either the target is detected correctly or no target is detected in a frame without a target). It is defined as $(N_{TP} + N_{TN}) / N_{All}$.
\item Precision measures what ratio of all the target detections is actually correct. It is defined as $N_{TP} / (N_{TP} + N_{FP} + N_{WD})$.
\item Recall or sensitivity measures the ratio of all the frames containing targets that are correctly detected. It is defined as $N_{TP} / (N_{TP} + N_{FN} + N_{WD})$.
\item F1 Score is the weighted average of precision and recall and takes both false positives and false negatives into account. F1 Score is defined as $2 \times (Recall \times Precision) / (Recall + Precision)$.
\end{itemize}

It can be seen that increasing the value of \textit{ContourOverlap} up to some point generally decreases the number of false results (both false positives and false negatives) and therefore increases the number of correct results and statistical measures. Although, after some point, the number of detections (true or false positives) starts dropping, and the performance slowly decreases. Additionally, the higher \textit{ContourOverlap} results in more candidate contours being eliminated before further processing, reducing the algorithm's execution time. 

On the other hand, the value of \textit{EllipseOverlap} has a smaller effect on performance and execution time, slightly improving the performance up to a point. If it is set too high, the algorithm starts rejecting more ellipses, causing a drop in the number of positive results, significantly decreasing the algorithm's performance. 

%Finally, on average, the frames other than True Positives require two detection iterations for each frame, making them slower than True Positive frames.

The choice of other parameters in the algorithm depends on the application. For example, when the pattern is not expected to be far, setting a higher number for $minContourSize$ will improve the speed by eliminating the contours that cannot belong to the elliptic pattern. On the other hand, setting this parameter too high can result in false negatives when the pattern is seen partially or consists of several partial contours. In practice, however, the method has shown to be rather insensitive to these parameters, and generally, there is no need to change the default values. In all our tests, we have used the following values: $mnAxSize = 5$, $mxAxSize = 700$, $maxAxisRatio = 5$, $minContourSize = 50$, $maxScale = 100$.

\subsection{Comparison With Other Methods} \label{sec:comparison}

To compare the performance of our algorithm with other methods, we ran four other methods on the dataset introduced in Section~\ref{sec:dataset}:
the MATLAB implementation of a Hough Transform-inspired approach proposed by Xie \& Ji~\cite{xie2002ellipsedetect} with a random sub-sampling inspired by the work in~\cite{basca2005htsampling}, 
and the C++ implementations of the methods proposed by Fornaciari et al.~\cite{fornaciari2014ellipsedetect}, Prasad et al.~\cite{prasad2012}, and Jia et al.~\cite{jia2017}.
Table~\ref{tbl:ellipse-methods-performance} shows the results and performance of the algorithms on our dataset.

\begin{table}[!t]
\centering
\begin{threeparttable}
\caption{Performance of the ellipse detector methods on the AirLab Elliptic Target Detection Dataset.}
\label{tbl:ellipse-methods-performance}
\begin{tabular}{|c|c|c|c|c|c|c|c|c|c|}
\hline
\rowcolor[HTML]{EFEFEF} 
Method            & Accuracy$^{\star}$ & Precision$^{\star}$ & Recall$^{\star}$ & F1 Score$^{\star}$ \\ \hline
Ours         & \textbf{96.56\%}                                               & \textbf{99.77\%}                                                & \textbf{96.44\%}                                                           & \textbf{0.981}    \\ \hline
\cite{xie2002ellipsedetect}  & 3.64\%                                                & 3.64\%                                                 & 3.99\%                                                            & 0.038    \\ \hline
\cite{fornaciari2014ellipsedetect}  & 88.75\%                                               & 99.67\%                                                & 87.66\%                                                           & 0.933    \\ \hline
\cite{prasad2012}  & 89.81\%                                               & 99.45\%                                                & 90.22\%                                                           & 0.965    \\ \hline
\cite{jia2017}  & 78.36\%                                               & 95.50\%                                                & 80.04\%                                                           & 0.871    \\ \hline
\end{tabular}
    \begin{tablenotes}
      \small
      \item $^{\star}$ As defined in Section~\ref{sec:dataset}.
    \end{tablenotes}
\end{threeparttable}
\end{table}

The results show that our implemented algorithm outperforms the other methods in all the criteria. The method by Xie \& Ji's~\cite{xie2002ellipsedetect} was unable to perform well on the real frames of our test environment due to the relatively low resolution of our frames and the small size of the target in the frames; the few cases it could detect the elliptic target were when the target was covering a large portion of the frame. The main problem with the method by Fornaciari~\cite{fornaciari2014ellipsedetect}  was that it was unable to detect the elliptic targets when more than 25\% of the target was outside of the frame (case (e) in Figure~\ref{fig:circletypes}). The method by Jia et al.~\cite{jia2017} is comparatively fast but has a high false positive rate and is unable to detect the ellipses when they are far away (small in the frame). Prasad's method~\cite{prasad2012} has comparable performance to Fornaciari's but is significantly slower. At the same time, our proposed algorithm was fast and still able to detect the target's elliptic pattern in partial views or when it was small. Additionally, we should note that all the false positive cases of our algorithm on the dataset, detected elliptical drawings on the ground, which would have been rejected if the UAV's altitude information was used.

%Comparing the above results with the ellipse detection results reported on a dataset gathered for similar patterns by \cite{jin2018}, we notice the similar F1~score reported for the \cite{fornaciari2014ellipsedetect} method, which can be an indicator of the similarity of the datasets and the testing conditions. While we recognize that the results of two methods on different datasets cannot be confidently used for the comparison of performance, we do believe that our method outperforms those methods as well since the F1~score of our method compared to all the five other methods in \cite{jin2018} is higher. We cannot claim the same for the execution times, as the hardware used for the tests has a significant effect on the results, and the comparison on different platforms will be meaningless. 

\begin{figure}[!t]
\centering
    \begin{subfigure}[b]{0.23\textwidth}
        \includegraphics[width=\textwidth]{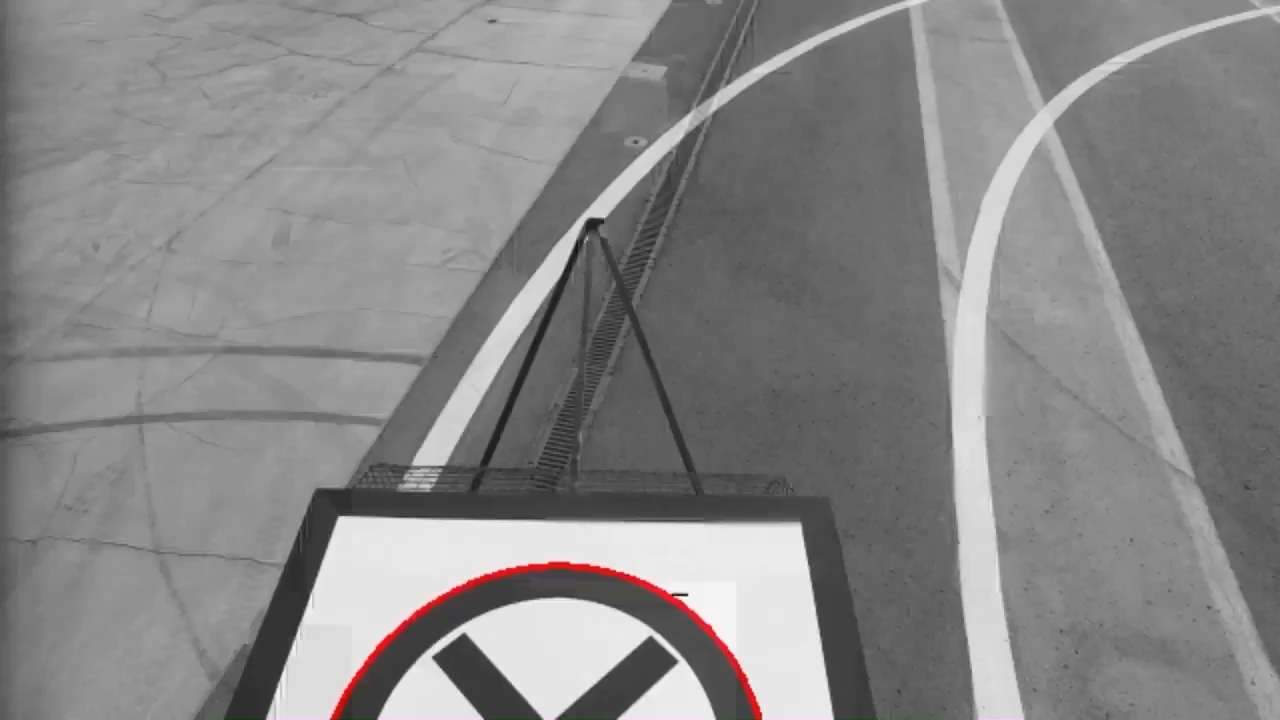}
        \caption{~}
    \end{subfigure}
    \hfill
    \begin{subfigure}[b]{0.23\textwidth}
        \includegraphics[width=\textwidth]{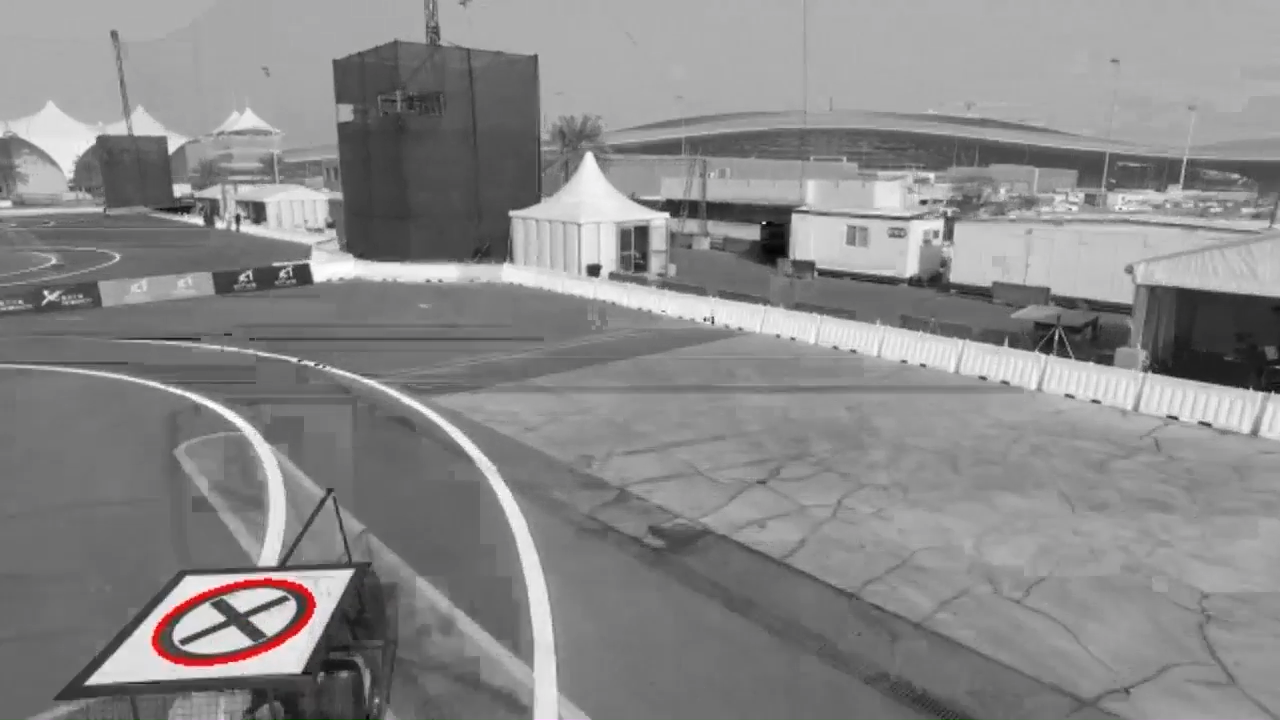}
        \caption{~}
    \end{subfigure}
    
    \medskip
    
    \begin{subfigure}[b]{0.23\textwidth}
        \includegraphics[width=\textwidth]{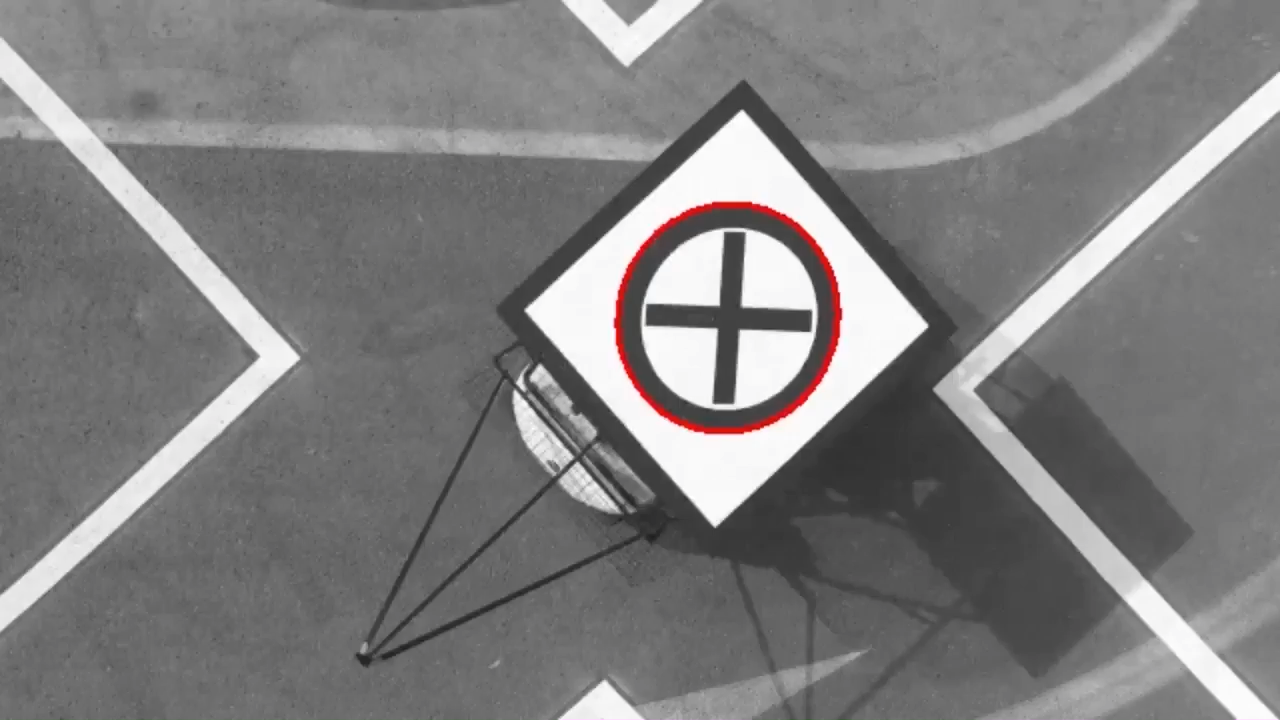}
        \caption{~}
    \end{subfigure}
    \hfill
    \begin{subfigure}[b]{0.23\textwidth}
        \includegraphics[width=\textwidth]{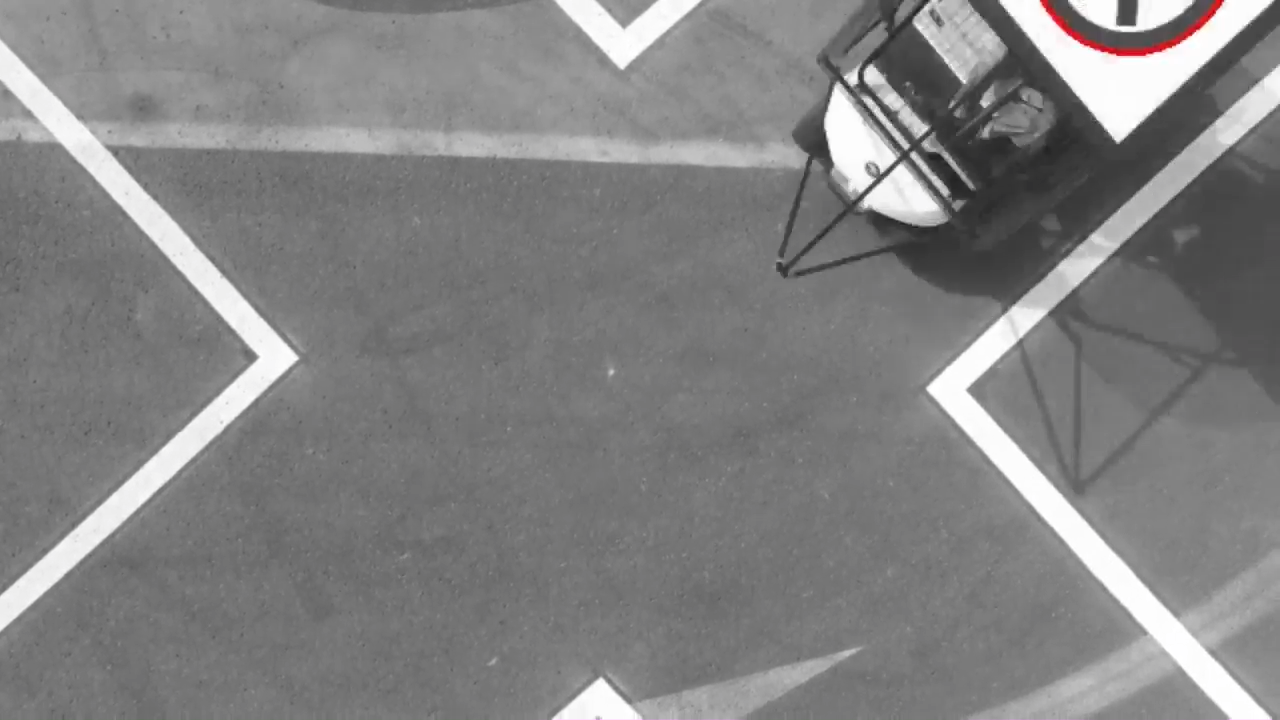}
        \caption{~}
    \end{subfigure}
\caption{Result of the proposed algorithm on sample frames. The red ellipse indicates the detected pattern on the deck of the moving vehicle.}
\label{fig:deckresults}
\end{figure}

Figure~\ref{fig:deckresults} shows results for the detection of the elliptical pattern in some sample frames from the dataset. The method by Fornaciari et al. is unable to detect the ellipses in Figure~\ref{fig:deckresults}(a) and Figure~\ref{fig:deckresults}(d). 

\begin{table*}[t]
\centering
\begin{threeparttable}
\caption{Execution times of ellipse detector methods on the AirLab Elliptic Target Detection Dataset.$^\star$}
\label{tbl:ellipse-methods-times}
\begin{tabular}{|c|c|c|c|c|c|c|c|c|c|c|c|c|}
\hline
\multirow{2}{*}{\textbf{Method}} & \multicolumn{3}{c|}{\textbf{True Positive}} & \multicolumn{3}{c|}{\textbf{True Negative}} & \multicolumn{3}{c|}{\textbf{False Positive}} & \multicolumn{3}{c|}{\textbf{False Negative}} \\ 
& \textit{Avg.}       & \textit{Max.}      & \textit{Min.}      & \textit{Avg.}      & \textit{Max.}       & \textit{Min.}      & \textit{Avg.}       & \textit{Max.}       & \textit{Min.}      & \textit{Avg.}       & \textit{Max.}       & \textit{Min.}      \\ \hline
Our Method              & 32.2    & \textbf{64.1}   & \textbf{1.5}    & 48.9   & 68.1   & \textbf{6.6}   & 48.4    & 49.6    & 46.4   & 39.1    & 64.0   & \textbf{2.5}    \\ \hline
Fornaciari, et al \cite{fornaciari2014ellipsedetect}%$^{\star\star}$
& 26.5    & 87.3   & 15.5   & 25.4   & 54.2    & 15.3   & -          & -          & -         & \textbf{23.8}    & \textbf{60.3}    & 14.7   \\ \hline
Prasad, et al \cite{prasad2012}%$^{\star\star}$
& 36.2    & 70.0   & 25.5   & 34.4   & 64.9    & 20.0   & -          & -          & -         & 38.8    & 72.1   & 26.6   \\ \hline
Jia, et al \cite{jia2017}%$^{\star\star}$
& \textbf{24.4}    & 67.3   & 13.1   & \textbf{21.2}   & \textbf{52.7}    & 11.6   & -          & -          & -         & \textbf{23.8}    & 63.3    & 12.8   \\ \hline
\end{tabular}
    \begin{tablenotes}
      \small
      \item $^\star$ All times are in milliseconds.
      %\item $^{\star\star}$ For this method, only the detection time is considered and the time of the target verification is not added to the result, which could potentially increase the times.
    \end{tablenotes}
\end{threeparttable}
\end{table*}

Table~\ref{tbl:ellipse-methods-times} compares the execution times of the ellipse detection methods on the same dataset (using Intel Core i5-4460 CPU @ 3.20 GHz). The implementation of Xie \& Ji's method is done in MATLAB, which gives much higher execution times than C++ implementations. Therefore we excluded it from Table~\ref{tbl:ellipse-methods-times}. Fornaciari et al. method has similar average execution times to our algorithm. However, their approach provides slightly more consistent execution times, which can be convenient for control systems using the detection output for robot control. On the other hand, our method's speed significantly increases (with frame processing time going down to just a few milliseconds) when the robot gets closer to the target pattern. This increase in speed especially helps the system to have a much higher detection rate when a higher processing speed is needed for the robot to approach the moving vehicle for landing.

\subsection{Example Application} \label{sec:application}

\begin{figure}[!t]
\centering
    \begin{subfigure}[b]{0.23\textwidth}
        \includegraphics[width=\textwidth]{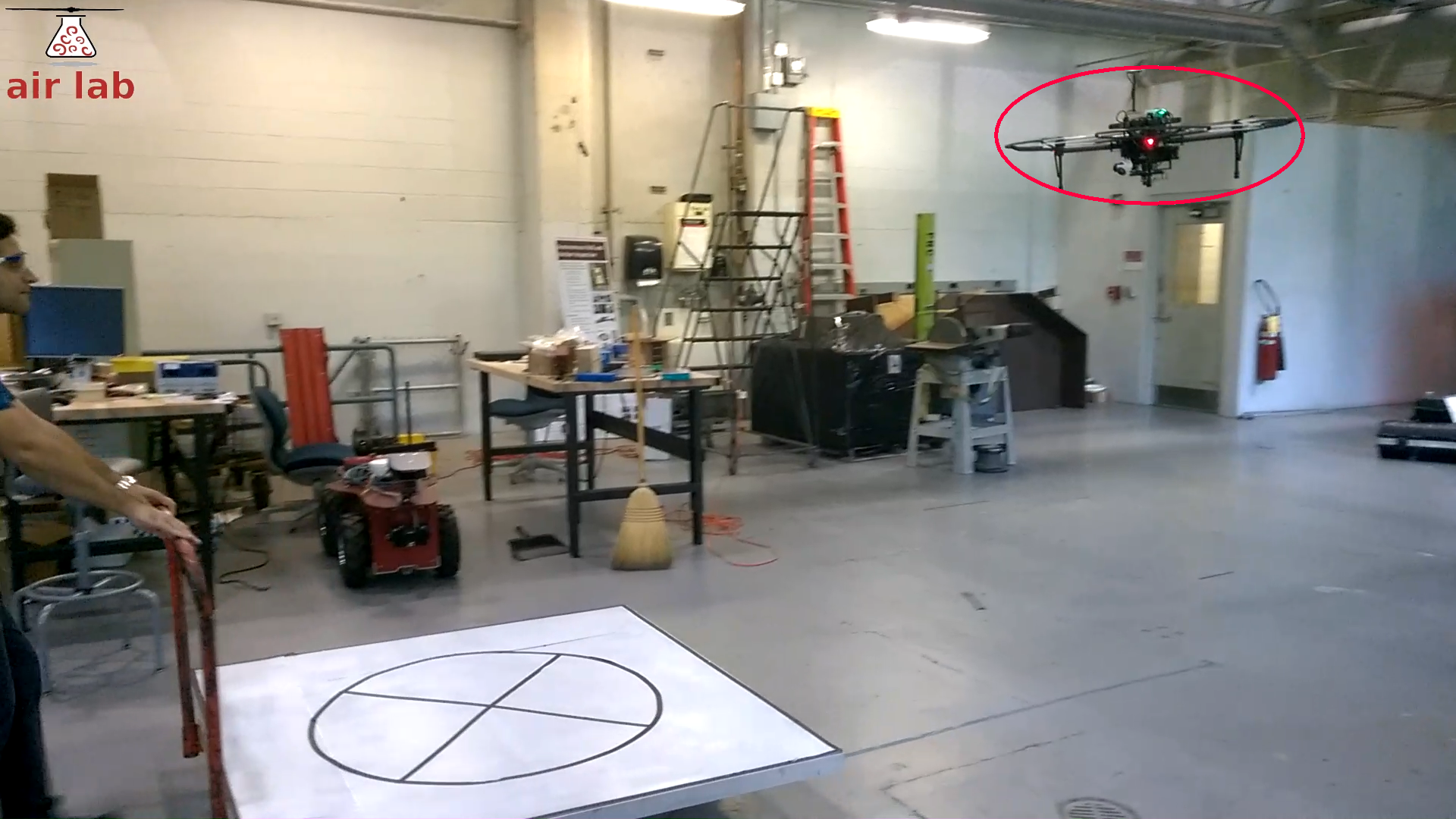}
        \caption{~}
    \end{subfigure}
    \hfill
    \begin{subfigure}[b]{0.23\textwidth}
        \includegraphics[width=\textwidth]{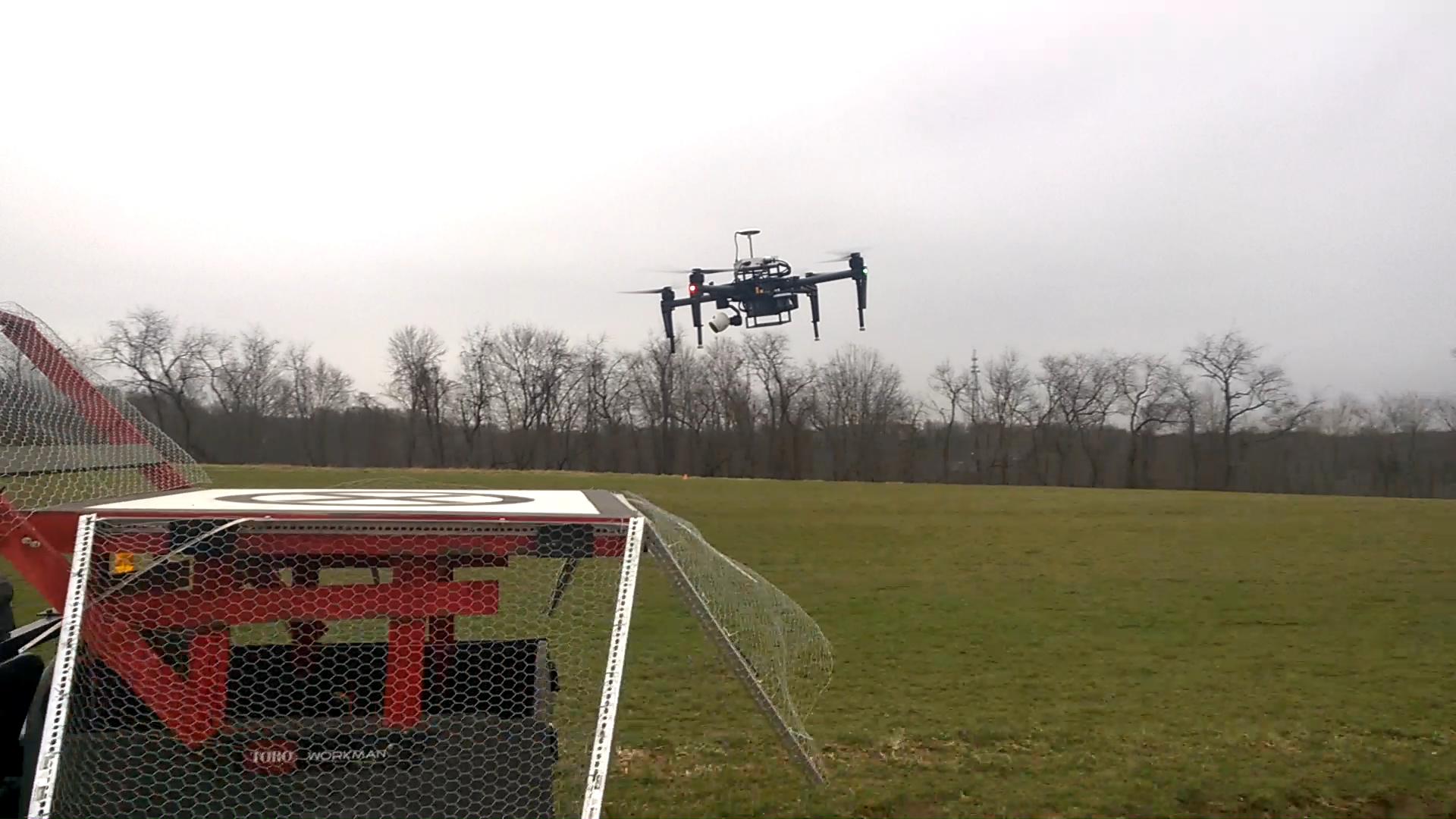}
        \caption{~}
    \end{subfigure}
    
    \medskip
    
    \begin{subfigure}[b]{0.23\textwidth}
        \includegraphics[width=\textwidth, height=2.4cm]{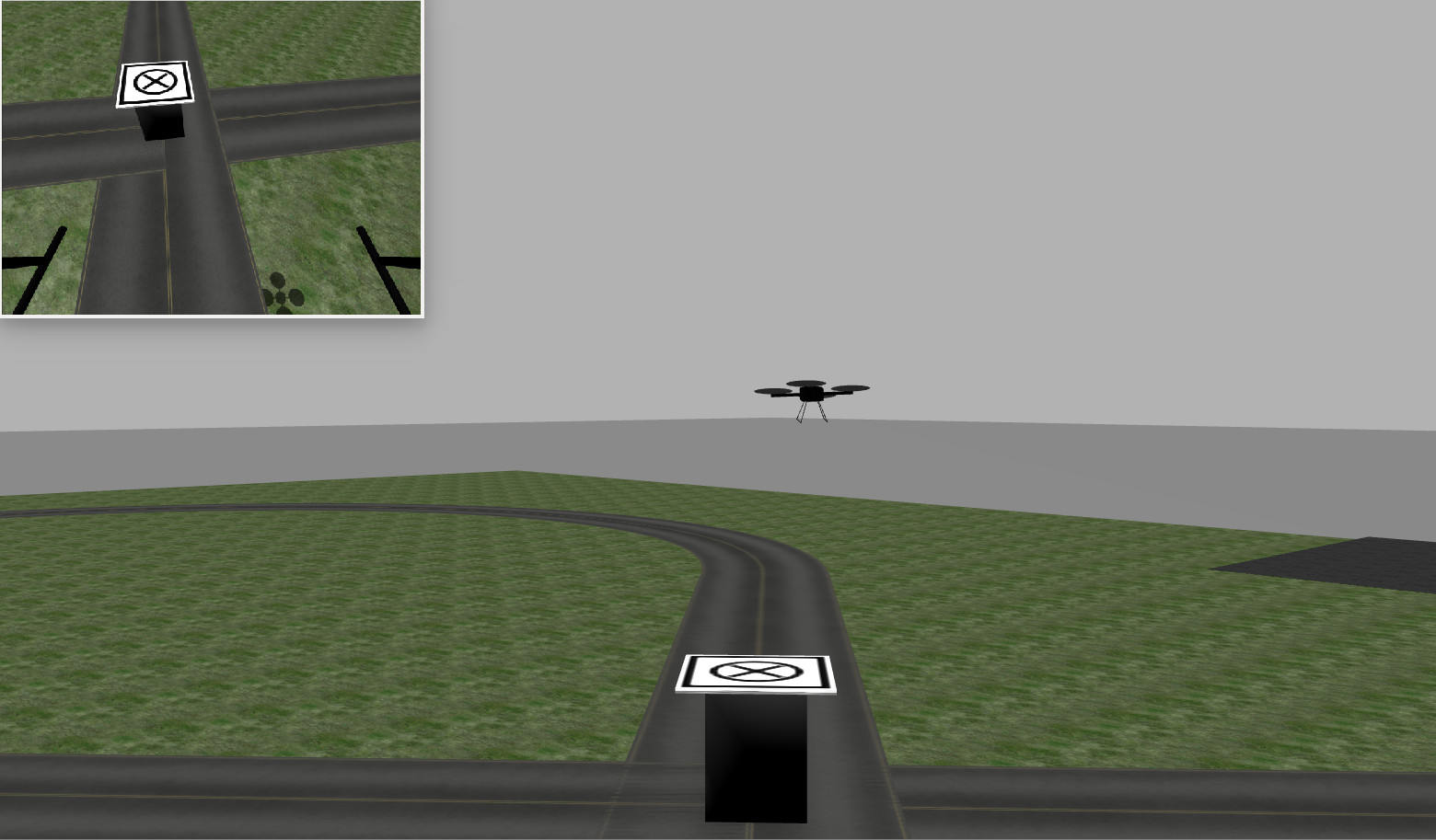}
        \caption{~}
    \end{subfigure}
    \hfill
    \begin{subfigure}[b]{0.23\textwidth}
        \includegraphics[width=\textwidth, height=2.4cm]{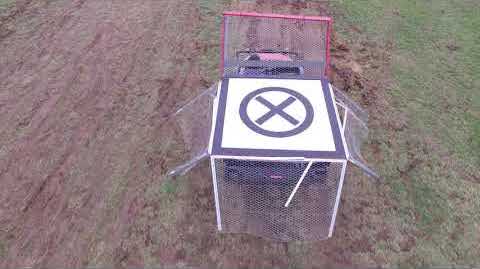}
        \caption{~}
    \end{subfigure}
    
\caption{Screenshots from video sequences showing our autonomous UAV landing on a pattern moving at up to $4.2$~m/s speed in various lighting and environmental conditions.}
\label{fig:landing-tests}
\end{figure}

To test the proposed ellipse detection and tracking method's performance, it was used with a visual servoing method for an autonomous UAV landing on a circular pattern painted on top of moving platforms in indoor, outdoor, and simulated environments~\cite{keipour2021visualservoing}. Figure~\ref{fig:landing-tests} shows screenshots of the method in these different lighting conditions. Videos for these experiments and the project details can be accessed at \url{http://theairlab.org/landing-on-vehicle}.

%% file: 4.Conclusion.tex
\section{FUTURE WORK AND DISCUSSION} \label{sec:conclusion}

The proposed ellipse detection and tracking algorithm has shown its performance in an example application and has outperformed other standard methods in our tests. However, the proposed method has mainly been tested on a set of elliptical patterns used in our current and prior research. Utilizing the method for other applications may require further testing and improvements. The following suggestions can further enhance the performance and increase the method's usability in real-world robotics applications. 

The underlying algorithms used in the ellipse detection steps are the most common methods already available in the OpenCV library. The choice has been made to allow fast implementation by the potential reader and convenience. If better performance is required, the whole method's execution speed and performance can be improved by replacing steps such as ellipse fitting with faster and better algorithms.

Finally, if the robot's camera is not perfectly rectified, it may lose track of the elliptic target at close distances where only a small portion of the pattern is visible at such a skewed angle that it causes the circle to be seen as non-elliptic in the camera. The problem exists for any ellipse detection algorithm but can be improved using a robust tracker (such as Kernelized or Discriminative correlation filters) instead of a detector to track the target ellipse when the robot's camera is too close to the elliptical pattern.